\documentclass[lettersize,journal]{IEEEtran}
\usepackage{amsmath,amsfonts}
\usepackage{algorithmic}
\usepackage{algorithm}
\usepackage{array}
\usepackage[caption=false,font=normalsize,labelfont=sf,textfont=sf]{subfig}
\usepackage{textcomp}
\usepackage{stfloats}
\usepackage{url}
\usepackage{verbatim}
\usepackage{graphicx}
\usepackage{booktabs}
\usepackage{cite}
\usepackage{hyperref}

\begin{document}

\title{KnifeHunter: Structured Local Representation Learning for Fine-Grained Knife Image Retrieval in Law Enforcement}

\author{Syed Sameed Husain, Eng-Jon~Ong, Stephen Simpson, Trevor Hamshere, Matt Turner and Miroslaw Bober
\thanks{S. Husain, E. Ong and M. Bober are with the Centre for Vision, Speech and Signal Processing, University of Surrey, Guildford GU27XH, UK. e-mail: sh0057@surrey.ac.uk, e.ong@surrey.ac.uk, m.bober@surrey.ac.uk. \\
S. Simpson, T. Hamshere, M. Turner are with Metropolitan Police UK, e-mail: Businesscrime.365group@met.police.uk}
}

\markboth{Journal of \LaTeX\ Class Files,~Vol.~14, No.~8, August~2021}%
{Shell \MakeLowercase{\textit{et al.}}: A Sample Article Using IEEEtran.cls for IEEE Journals}


\maketitle

\begin{abstract}
Knife-enabled violence presents a major public safety challenge, and law enforcement agencies require scalable tools for catalogue-level knife identification, intelligence analysis, and source attribution. Manual visual comparison is specialist, time-consuming, and difficult to scale under operational imaging conditions. We introduce KnifeHunter, an end-to-end forensic knife image retrieval system developed with UK law enforcement. The work contributes the KnifeHunter dataset, comprising 25{,}843 images across 543 knife classes from police evidence, retail catalogues, and border-force seizures, with structured metadata, Medium/Hard evaluation protocols, and large-scale distractor evaluation. We further propose CoRe-Net, a compact single-descriptor retrieval architecture that combines global context with spatially localised discriminative evidence. CoRe-Net introduces Structured Complementary Representation Learning (SCRL) to organise local evidence into complementary prototype-based representations, and Bi-Directional Reciprocal Fusion (BDRF) to integrate global and local evidence through residual projection and gated local-to-global injection. Using an EVA02-Base backbone and cosine-similarity retrieval, CoRe-Net achieves 88.0\% mAP and 86.7\% mP@10 on the Medium protocol, and 85.1\% mAP and 83.8\% mP@10 under distractor conditions. 
KnifeHunter was deployed by UK police forces during Operation Sceptre deployments from 2023 to 2025, achieving 99.2\% mP@1 on field queries. These results demonstrate a practical and effective multimedia retrieval framework for fine-grained forensic knife matching in operational law-enforcement settings.

\end{abstract}


\section{Introduction}
\label{sec:intro}

\IEEEPARstart{K}{nife-enabled} violence is a major public safety concern in the United Kingdom and internationally. In the year ending March 2023, England and Wales recorded approximately 50,500 offences involving a knife or sharp instrument, motivating scalable tools for prevention, investigation, and intelligence-led policing. Recovered knives often need to be compared against reference catalogues to determine model, manufacturer, likely source, and legal classification. However, current workflows rely heavily on manual inspection and unstructured records, limiting scalability, cross-case analysis, and longitudinal intelligence gathering.

Knife identification presents a demanding fine-grained retrieval problem. Knives are widely available through lawful retail channels and illicit supply routes, may be modified after manufacture, and can differ only in subtle attributes such as blade geometry, serration pattern, locking mechanism, maker marks, or handle construction. At the same time, operational imagery is highly variable, with uncontrolled illumination, specular reflections, police rulers, evidence labels, occlusion, packaging, and cluttered backgrounds. These factors make catalogue-level matching difficult and require visual representations that preserve both global shape and spatially localised discriminative evidence.

\paragraph{Challenges in Intelligence Gathering}
Despite the value of knife intelligence, knives recovered during routine policing are often archived with limited structured information. Outside major investigations, records commonly consist of short free-text descriptions and may omit standardised imagery and searchable metadata. This restricts the ability to identify recurring models, connect seizures to common sources, quantify temporal or geographic trends, and coordinate analysis across police forces. Manual logging can enrich records, but it is difficult to sustain at operational scale.

\paragraph{The Computer Vision Perspective}
From a computer vision perspective, knife identification can be formulated as fine-grained instance-level image retrieval under adverse acquisition conditions. Unlike standard landmark or product retrieval, forensic knife imagery combines subtle inter-class differences, substantial intra-class variation from wear and damage, partial visibility, background clutter, and large distractor databases. Spatially sparse cues such as serrations, grind lines, tip shape, and maker marks may be decisive but occupy only limited image regions.

These conditions expose limitations in existing descriptor aggregation methods. Global pooling approaches such as GeM~\cite{gem} and SuperGlobal~\cite{supg} provide efficient single-descriptor retrieval but can be biased by background-driven activations and high-magnitude outliers. Local-global methods such as DOLG~\cite{dolg} and DELG~\cite{delg} improve representation capacity, but typically do not explicitly organise local evidence into complementary components or enforce reciprocal interaction between global and local representations within a compact embedding. This motivates a dedicated benchmark and aggregation strategy for preserving structured local evidence under forensic acquisition conditions.

\paragraph{Proposed Solution}
We introduce KnifeHunter, an AI-based system for fine-grained knife image retrieval, catalogue matching, and structured intelligence generation in law enforcement settings. Developed with the Metropolitan Police, the system enables officers to capture a field image and retrieve visually consistent candidates from a reference database together with catalogue metadata, transforming routine knife encounters into searchable intelligence for investigation, analysis, and enforcement.

To support this system, we propose CoRe-Net, a single-descriptor retrieval architecture that integrates complementary global and local evidence through structured descriptor aggregation and bidirectional fusion. The main contributions are:
\begin{enumerate}
    \item \textbf{KnifeHunter Dataset:} We introduce a forensic knife retrieval benchmark comprising 25{,}843 images across 543 catalogued knife classes, collected from police evidence, retail catalogues, and border-force seizures, with structured metadata for forensic search and analysis.
    
    \item \textbf{Complementary Representation Network (CoRe-Net):} We propose a retrieval architecture tailored to forensic imagery. CoRe-Net introduces Structured Complementary Representation Learning (SCRL), which learns multiple complementary prototype-based representations of spatially localised discriminative evidence, and Bi-Directional Reciprocal Fusion (BDRF), which couples global and local representations through residual projection and gated local-to-global injection. The resulting compact descriptor supports efficient cosine-similarity retrieval.
    
    \item \textbf{Benchmarking and Operational Validation:} We evaluate modern backbones and state-of-the-art aggregation modules under Medium and Hard protocols, with and without distractors, and report deployment experience with UK law enforcement during \textit{Operation Sceptre}.
\end{enumerate}

\section{Related Work}
\label{sec:related_work}

This work relates to instance-level image retrieval, compact descriptor aggregation, and computer vision for forensic and law-enforcement applications.

\subsection{Instance-Level Image Retrieval and Datasets}

Instance-level image retrieval aims to recover images depicting the same physical object under variations in viewpoint, illumination, occlusion, and background. Benchmarks such as Revisited Oxford and Paris~\cite{oxf}, ILIAS~\cite{ilias}, and Google Landmarks v2~\cite{google} have advanced retrieval research through standardised protocols, large-scale training, and distractor-based evaluation. However, these datasets primarily target landmarks or broad consumer objects. In contrast, KnifeHunter focuses on forensic knife retrieval, combining police evidence imagery, retail catalogue images, border-force seizures, structured metadata, and operationally motivated evaluation protocols.

\subsection{Descriptor Learning for Image Retrieval}

Single-descriptor retrieval remains attractive for large-scale search because it supports efficient nearest-neighbour indexing. GeM~\cite{gem} learns a pooling exponent that interpolates between average and max pooling, while SuperGlobal~\cite{supg} improves global-descriptor retrieval through GeM+, Regional-GeM, Scale-GeM, and global-descriptor re-ranking. Although effective, global aggregation can be biased by background responses or high-magnitude activations and may under-represent discriminative evidence confined to small image regions.

Several methods address this limitation by incorporating local or structural information. TokenNet~\cite{token} aggregates dense spatial features into attention-weighted tokens and refines them through self- and cross-attention. DOLG~\cite{dolg} performs orthogonal fusion between local and global representations to reduce redundancy, while DELG~\cite{delg} combines a compact global descriptor with attention-guided local features and optional spatial verification. SENet~\cite{senet} incorporates structural self-similarity by encoding local self-similarity descriptors and fusing them with appearance features before global pooling. Re-ranking and verification approaches, including local feature matching, correlation verification~\cite{cvnet}, and asymmetric similarity estimation~\cite{ames}, can further improve accuracy but introduce additional computational cost and complexity.

CoRe-Net differs from these methods by explicitly organising spatially localised evidence into complementary prototype-based representations and coupling them with the global descriptor through reciprocal fusion. This preserves single-descriptor retrieval without query expansion, re-ranking, or geometric verification, which is important for deterministic and efficient forensic deployment.

\subsection{Computer Vision for Forensic Applications}

Computer vision has been applied to forensic and law-enforcement tasks including biometric identification~\cite{lin2018fingerprint}, firearm and toolmark analysis~\cite{giverts2024firearm, cuellar2024toolmark}, shoeprint retrieval and comparison~\cite{cui2019shoeprint,wen2023shoeprint}, and weapon detection in surveillance imagery~\cite{shanthi2025weapon}. These works typically address controlled biometric matching, forensic comparison, or detection-oriented objectives. KnifeHunter addresses a different setting by formulating knife cataloguing as fine-grained instance-level retrieval under operational evidence imagery and open-source images, contributing both a benchmark with structured protocols and a retrieval architecture designed for cluttered and distractor-rich search.

\section{KnifeHunter Dataset}
\label{sec:dataset}

A central contribution of this work is the KnifeHunter dataset, a benchmark designed for forensic knife image retrieval. The dataset is constructed to reflect operational evidence imagery and deployment conditions encountered in law enforcement workflows, including visually similar knife classes, variable scene contexts, and retrieval in the presence of large-scale distractor sets.

\subsection{Dataset Composition}

\begin{figure*}[t]
    \centering
    \includegraphics[width=0.8\linewidth]{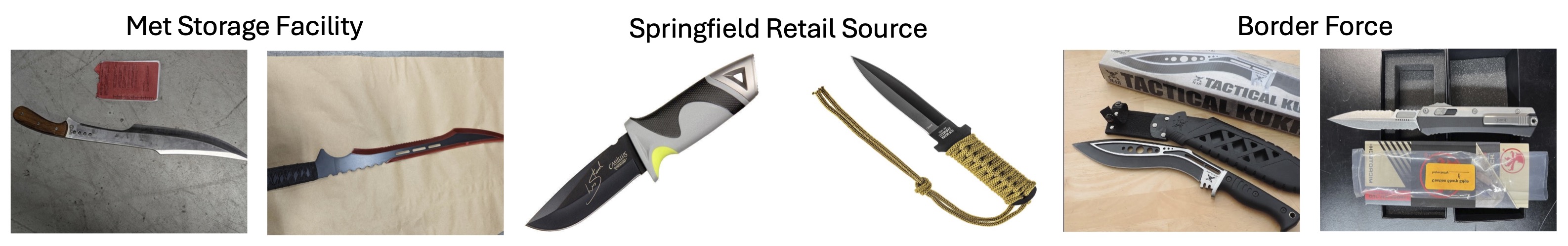}
    \caption{Representative knife images from the Metropolitan Police storage facility, Springfields retail listings, and UK Border Force seizures.}
    \label{fig:knives}
\end{figure*}

The KnifeHunter dataset comprises 25{,}843 images across 543 distinct knife classes collected from three complementary sources: police evidence repositories, public retail listings, and border-force seizure records. Together, these sources capture knives encountered in criminal investigations, legal retail markets, and prohibited importation scenarios. Representative samples are shown in Fig.~\ref{fig:knives}.

The police evidence subset is the largest and most operationally representative component, comprising 332 knife classes photographed at Metropolitan Police storage facilities. Images were captured with smartphones and DSLR cameras under varied viewpoints, illumination, backgrounds, and evidence-documentation conditions, including police rulers, evidence bags, sheaths, and forensic tools. Additional acquisition examples are provided in Supplementary Section A. The retail subset contains 171 knife classes from Springfields product listings and was augmented using geometric and photometric transformations to improve visual diversity. The UK Border Force subset contains 40 knife classes from customs seizures, including uncommon, modified, or prohibited imports.

\subsection{Evaluation Dataset and Protocol}

\begin{figure*}[t]
    \centering
    \includegraphics[width=0.8\linewidth]{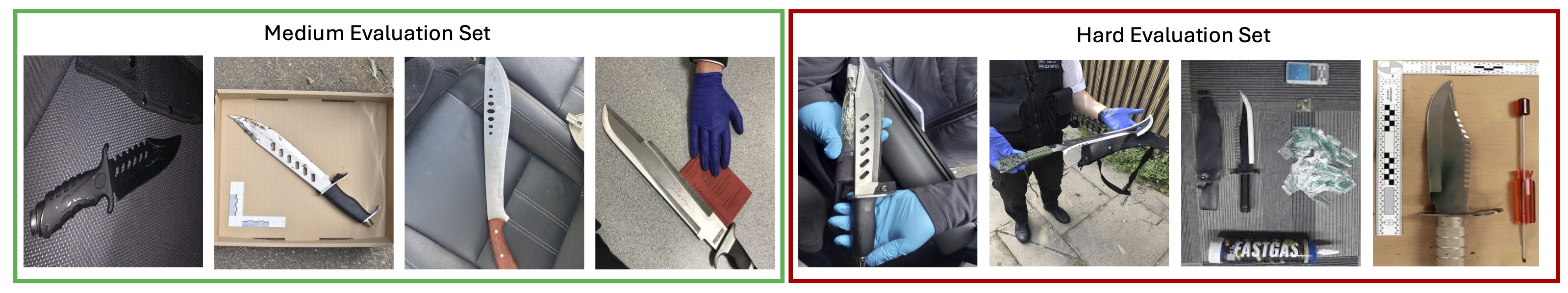}
    \caption{Representative samples from the Medium and Hard evaluation query subsets.}
    \label{fig:eval}
\end{figure*}

We define two evaluation query subsets of increasing difficulty. The \textbf{Medium} set contains 443 query images captured under moderately challenging field conditions, including viewpoint variation, partial occlusion, background clutter, and illumination changes. The \textbf{Hard} set contains 150 query images sourced predominantly from knife-crime investigations and includes severe viewpoint changes, heavy occlusion, cluttered scenes, poor illumination, and non-canonical orientations. Representative queries are shown in Fig.~\ref{fig:eval}.

To simulate realistic deployment, we also evaluate retrieval with 1{,}131{,}594 distractor images. These comprise approximately 1M standard retrieval distractors~\cite{oxf}, 10{,}000 knife-related images from UK police forces that are disjoint from the KnifeHunter training and test classes, and additional images sampled from COCO 2017 Unlabelled. This setting introduces both visually relevant unseen-knife distractors and visually irrelevant background distractors, requiring models to remain selective at a large scale.

\subsection{Evaluation Metrics}

At inference, each query and database image is encoded as a single descriptor, $\ell_2$-normalised, and ranked by cosine similarity. We report mean Average Precision (mAP)~\cite{oxf} to measure overall ranking quality and mean Precision at rank 10 (mP@10) to evaluate the top results inspected by analysts. Metrics are reported on the Medium and Hard protocols, with and without the 1.13M distractor database.

\begin{figure*}[t]
    \centering
    \begin{tabular}{@{}c@{\hspace{0.03\linewidth}}c@{}}
        \includegraphics[width=0.4\linewidth]{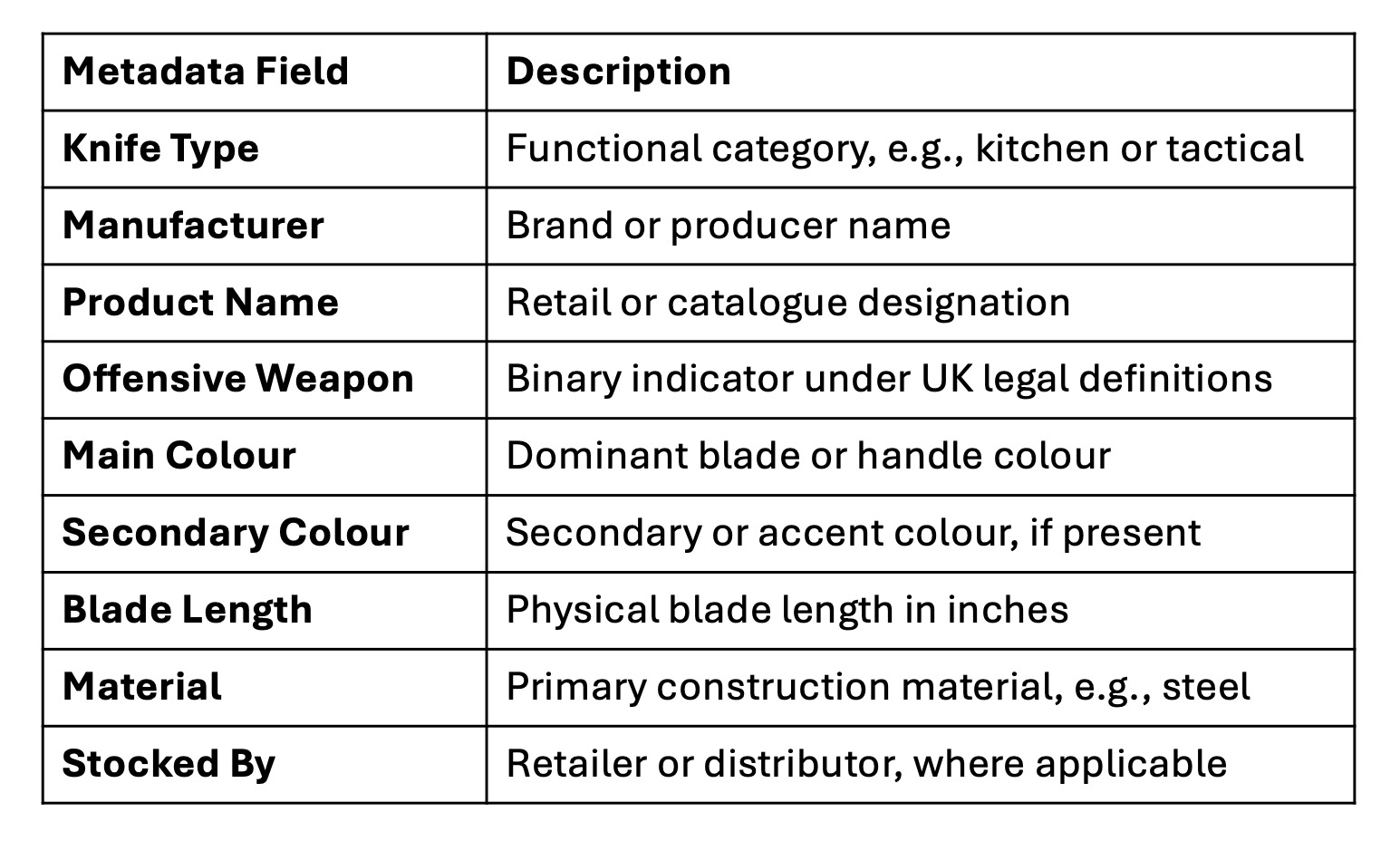} &
        \includegraphics[width=0.4\linewidth]{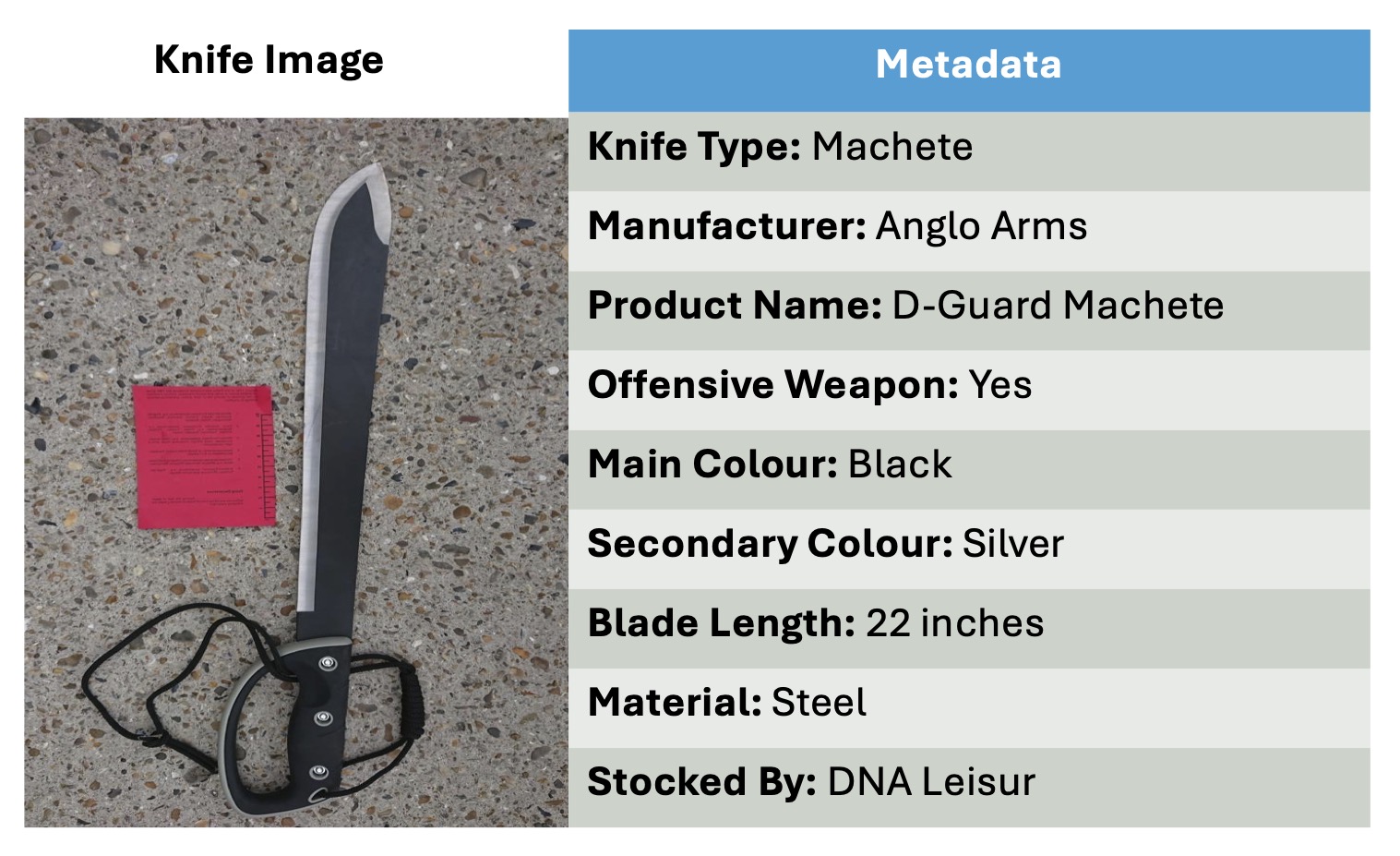} \\
        \textbf{(a)} & \textbf{(b)}
    \end{tabular}
    \caption{(a) Metadata fields associated with each knife instance. 
    (b) Example knife instance with its corresponding metadata record.}
    \label{fig:metadata_combined}
\end{figure*}

\subsection{Metadata and Benchmark Significance}
\label{sec:metadata_benchmark}

Each knife instance is annotated with structured metadata developed with the Metropolitan Police and informed by consultations with UK police forces, manufacturers, and retailers. The metadata supports forensic search, filtering, catalogue matching, and legal analysis. Fig.~\ref{fig:metadata_combined} summarises the metadata fields and provides an example record.

KnifeHunter provides a realistic benchmark for forensic object retrieval by combining heterogeneous data sources, structured metadata, difficulty-based evaluation splits, and large-scale distractor evaluation. These design choices reflect operational law-enforcement constraints and support rigorous evaluation of retrieval architectures intended for deployment.

\section{Complementary Representation Network}
\label{sec:method}

We propose the Complementary Representation Network (CoRe-Net), a single-descriptor architecture for fine-grained instance-level knife retrieval under forensic imaging conditions. Given an input image, CoRe-Net produces a compact embedding that is compared using cosine similarity for nearest-neighbour search in large retrieval databases.

CoRe-Net is designed to address two key limitations of existing aggregation methods: the difficulty of preserving multiple spatially localised sources of discriminative evidence present in the image within a compact descriptor, and the lack of mechanisms to enforce complementarity between global context and local structure during descriptor formation. To address these challenges, CoRe-Net introduces a local aggregation mechanism that organises features into multiple complementary prototype-based representations, together with a bidirectional fusion mechanism that explicitly couples the global context and local structure within a single compact descriptor. 

\begin{figure*}[t]
    \centering
    \includegraphics[width=0.8\linewidth]{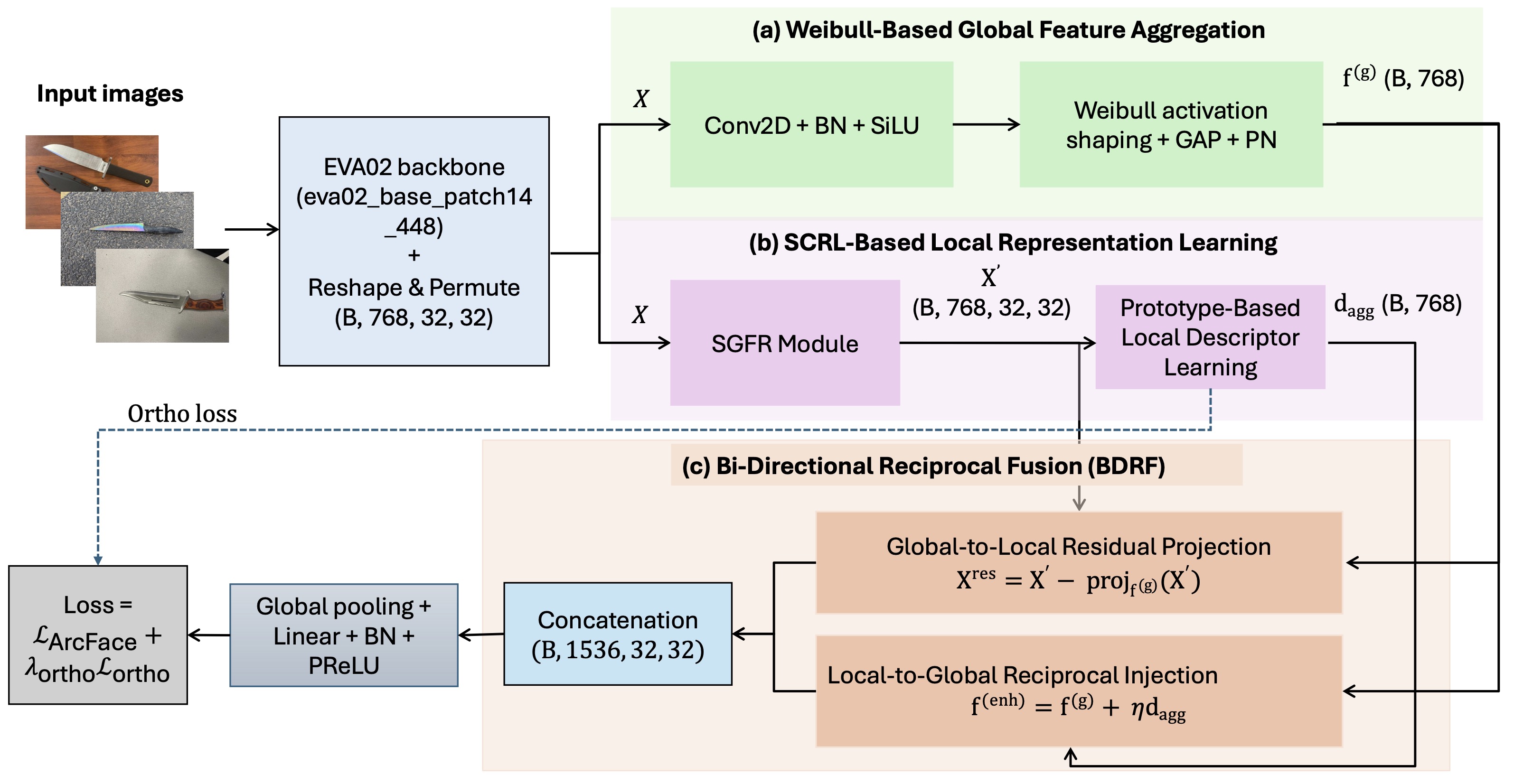}
    \caption{Proposed Complementary Representation Network (CoRe-Net). A backbone feature tensor is processed by a Weibull-based global aggregation branch and an SCRL local representation branch. SCRL performs saliency-guided refinement and prototype-based local descriptor learning, while BDRF integrates global and local evidence through residual projection and gated local-to-global injection. The fused representation is pooled and projected into a 512-dimensional embedding for retrieval.}
    \label{fig:arch}
\end{figure*}

As shown in Fig.~\ref{fig:arch}, the proposed architecture consists of three main components: (a) a global aggregation branch that produces a robust global descriptor, (b) a local aggregation branch based on Structured Complementary Representation Learning (SCRL), and (c) a Bi-Directional Reciprocal Fusion (BDRF) module that integrates global and local information. Among these, SCRL and BDRF constitute the primary methodological contributions of this work, while the remaining components support stable and effective descriptor learning.

\subsection{Backbone Feature Extraction}
Let an input image be processed by a backbone network to yield a dense feature tensor
\begin{equation}
\mathbf{X}\in\mathbb{R}^{K\times H\times W},
\end{equation}
where $K$ is the channel dimension and $H\times W$ denotes spatial resolution. For transformer backbones, patch tokens are reshaped into a spatial grid after removing special tokens. We denote the activation at channel $k$ and location $(h,w)$ as $X_{k,h,w}$.

\subsection{Weibull-based Global Representation}
\label{sec:weibull_global}

Knife retrieval relies heavily on global appearance information, but forensic evidence imagery often contains structured clutter and uncontrolled illumination, which can produce numerous weak background activations and a small number of very strong responses from specular highlights or unrelated structures. To reduce the influence of both effects, CoRe-Net applies a learnable Weibull-based activation shaping function before global average pooling and power normalisation. The shaping is applied to an adapted feature map produced by a lightweight convolutional block:
\begin{equation}
\tilde{\mathbf{X}} = \phi\!\left(\mathrm{BN}\!\left(\mathrm{Conv}_{3\times3}(\mathbf{X})\right)\right),
\label{eq:global_adapt}
\end{equation}
where \(\phi(\cdot)\) denotes the SiLU activation. Since the Weibull formulation requires non-negative inputs, we rectify the adapted feature map by clamping:
\begin{equation}
\bar{X}_{k,h,w}=\max(\tilde{X}_{k,h,w},\epsilon),
\label{eq:clamp}
\end{equation}
with a small constant \(\epsilon>0\).

For an activation \(\bar{x}=\bar{X}_{k,h,w}\), the Weibull-based activation shaping is defined as
\begin{equation}
\theta(\bar{x})=\left(\frac{\bar{x}}{\alpha}\right)^{\beta-1}
\exp\!\left(-\left(\frac{\bar{x}}{\gamma}\right)^{\zeta}\right),
\label{eq:weibull_activation}
\end{equation}
where \(\alpha,\beta,\gamma,\zeta\) are learnable scalar parameters shared across channels and spatial locations. The Weibull transformation combines polynomial growth with exponential decay. This reduces the influence of weak background responses and prevents a small number of high-magnitude activations from dominating the pooled descriptor. The shaped activations are aggregated using global average pooling:
\begin{equation}
f^{(g)}_k = \frac{1}{HW}\sum_{h=1}^{H}\sum_{w=1}^{W}\theta\!\left(\bar{X}_{k,h,w}\right),
\label{eq:global_pool_scalar}
\end{equation}
\begin{equation}
\mathbf{f}^{(g)} = \left[f^{(g)}_1,\dots,f^{(g)}_K\right]^{\top}\in\mathbb{R}^{K}.
\label{eq:global_pool_vec}
\end{equation}

A power-normalisation stage is then applied to stabilise the descriptor distribution and reduce sensitivity to residual outliers \cite{actnet}. Specifically, we apply an element-wise power transform with a learnable exponent \(\rho\) (initialised to \(\rho=1/3\)) to \(\mathbf{f}^{(g)}\), followed by \(\ell_2\) normalisation prior to fusion.

\subsection{Structured Complementary Representation Learning}
\label{sec:SCRL}

Global descriptors alone can be insufficient for separating visually similar knife classes when class separating evidence is confined to limited spatial regions, such as serration geometry, grind lines, tip shape, or maker marks. CoRe-Net introduces Structured Complementary Representation Learning (SCRL), a local aggregation mechanism that preserves and organises spatially localised discriminative evidence before fusion with the global branch.

SCRL consists of saliency-guided feature refinement, which attenuates locally homogeneous regions and increases the contribution of high-variance structures, and prototype-based local descriptor learning, which organises the refined local features into multiple complementary prototype-based representations. Each prototype acts as a learned representative of recurring local feature patterns.

\subsubsection{Saliency-Guided Feature Refinement (SGFR)}
Given the backbone feature map \(\mathbf{X}\), we compute a per-channel saliency mask from local activation variance within a \(3\times 3\) neighbourhood \(\mathcal{N}(h,w)\):
\begin{align}
\mu_{k,h,w} &= \frac{1}{|\mathcal{N}|}\sum_{(i,j)\in\mathcal{N}(h,w)} X_{k,i,j},\\
v_{k,h,w} &= \frac{1}{|\mathcal{N}|}\sum_{(i,j)\in\mathcal{N}(h,w)} X_{k,i,j}^{2} - \mu_{k,h,w}^{2}.
\end{align}
To ensure non-negative variance estimates, the raw variance is rectified as
\begin{equation}
\sigma^{2}_{k,h,w} = \max(v_{k,h,w},0) + \epsilon,
\end{equation}
where \(\epsilon>0\) is a small constant. The rectified variance is then batch normalised and mapped to a gating mask through a sigmoid function with learnable scale and bias parameters:

\begin{equation}
M_{k,h,w}=\operatorname{sigmoid}\!\Big(\alpha_k\,\mathrm{BN}(\sigma^2_{k,h,w})+\beta_k\Big),
\label{eq:sgfr_mask}
\end{equation}
with learnable per-channel parameters \(\{\alpha_k,\beta_k\}_{k=1}^{K}\). The refined feature map is
\begin{equation}
\mathbf{X}'=\mathbf{X}\odot\mathbf{M}.
\label{eq:xprime}
\end{equation}

This operation suppresses regions with weak local variation, such as uniform background surfaces and saturated glare, and increases the contribution of spatial locations with stronger local structural variation. In forensic knife retrieval, these regions are more likely to contain discriminative knife evidence.

\subsubsection{Prototype-Based Local Descriptor Learning (PBLD)}
Direct aggregation of the refined feature map \(\mathbf{X}'\) to a single descriptor may under-represent weak but discriminative evidence that occupies limited spatial support. CoRe-Net therefore introduces a small set of learnable prototypes that provide a compact latent basis for recurring local appearance patterns. This enables the model to organise the refined local features into multiple discriminative components before forming the injected local descriptor.

We reshape \(\mathbf{X}'\) into \(N=HW\) tokens \(\{\mathbf{t}_n\}_{n=1}^{N}\), where \(\mathbf{t}_n\in\mathbb{R}^{K}\) and learn a prototype matrix
\[
\mathbf{P}=[\mathbf{p}_1,\dots,\mathbf{p}_P]^\top \in \mathbb{R}^{P\times K},
\]
with \(\mathbf{p}_p\in\mathbb{R}^{K}\). In the implementation, the prototypes are initialised as trainable parameters drawn from a zero mean normal distribution and are optimised jointly with the backbone and fusion modules through end-to-end training. No semantic labels are imposed on individual prototypes; instead, they are learned as latent local basis vectors adapted to the retrieval objective.

Soft assignments between tokens and prototypes are computed using scaled dot-product attention:
\begin{equation}
a_{n,p}
=
\frac{\exp\!\left(\mathbf{t}_n^\top\mathbf{p}_p/\sqrt{K}\right)}
{\sum_{q=1}^{P}\exp\!\left(\mathbf{t}_n^\top\mathbf{p}_q/\sqrt{K}\right)}.
\label{eq:assign}
\end{equation}
This assignment mechanism allows each token to contribute to multiple prototypes with different weights, which is preferable to hard assignment when knife evidence is incomplete, spatially fragmented, or mixed with background content.

Prototype descriptors are then obtained by attention weighted aggregation:
\begin{equation}
\mathbf{d}_p = \sum_{n=1}^{N} a_{n,p}\,\mathbf{t}_n,
\qquad
\mathbf{D}=[\mathbf{d}_1,\dots,\mathbf{d}_P]^\top\in\mathbb{R}^{P\times K},
\label{eq:proto_desc}
\end{equation}
followed by layer normalisation applied independently to each \(\mathbf{d}_p\). This yields a compact and order invariant representation of the refined local map, in which different prototype descriptors can adapt to different local appearance patterns. Compared with the direct aggregation of \(\mathbf{X}'\) to a single descriptor, this representation retains greater local discrimination while preserving a fixed-dimensional summary suitable for single descriptor retrieval. 

\subsubsection{Prototype Diversity via Orthogonality Regularisation}
To reduce prototype collapse and encourage diversity, the prototypes are \(\ell_2\) normalised and their Gram matrix is regularised toward the identity. Defining the normalised prototypes as
\begin{equation}
\hat{\mathbf{p}}_p=\frac{\mathbf{p}_p}{\|\mathbf{p}_p\|_2+\epsilon},\qquad
\hat{\mathbf{P}}=[\hat{\mathbf{p}}_1,\dots,\hat{\mathbf{p}}_P]^\top\in\mathbb{R}^{P\times K},
\end{equation}
the corresponding Gram matrix is
\begin{equation}
\mathbf{G}=\hat{\mathbf{P}}\hat{\mathbf{P}}^\top\in\mathbb{R}^{P\times P}.
\end{equation}
The orthogonality regulariser is defined as
\begin{equation}
\mathcal{L}_{\mathrm{ortho}}=\frac{1}{P^2}\sum_{i=1}^{P}\sum_{j=1}^{P}\left(G_{ij}-I_{ij}\right)^2,
\label{eq:ortho}
\end{equation}
which corresponds to the element-wise mean squared deviation of \(\mathbf{G}\) from the identity matrix. This regularisation encourages different prototypes to encode distinct and complementary patterns of spatially localised discriminative evidence, reducing redundancy while increasing the diversity of cues preserved in the final representation.

\subsubsection{Prototype Gated Aggregation}
To form a compact representation of the refined local features, CoRe-Net computes a learned mixture over the prototype descriptors.
Each prototype descriptor \(\mathbf{d}_p\in\mathbb{R}^{K}\) is passed through a two-layer multilayer perceptron comprising a linear projection from \(K\) to \(K/2\) dimensions, a ReLU nonlinearity, and a second linear projection from \(K/2\) dimensions to a scalar logit \(g_p\). A softmax is then applied across the \(P\) prototypes:
\begin{equation}
\omega_p=\frac{\exp(g_p)}{\sum_{q=1}^{P}\exp(g_q)},
\qquad
\mathbf{d}_{\mathrm{agg}}=\sum_{p=1}^{P}\omega_p\,\mathbf{d}_p
\in\mathbb{R}^{K}.
\label{eq:dagg}
\end{equation}
This yields a weighted average of the prototype descriptors, where the weights are adapted to the input image. The softmax normalisation ensures that the weights are non negative and sum to one.

\subsection{Bi-Directional Reciprocal Fusion (BDRF)}
\label{sec:bdrf}

CoRe-Net fuses global and local information using a reciprocal fusion mechanism that explicitly promotes complementarity between the two branches. The global branch produces a descriptor \(\mathbf{f}^{(g)}\) capturing coarse knife geometry and overall appearance, while the local branch yields both a refined local feature map \(\mathbf{X}'\) and an aggregated prototype descriptor \(\mathbf{d}_{\mathrm{agg}}\). Thus, at the fusion stage, the model combines the global descriptor with both a spatially resolved local representation and a compact prototype-aggregated local summary.

BDRF addresses two complementary objectives. First, it removes from each local feature the component aligned with the global descriptor, encouraging it to focus on the residual local evidence. Second, it uses the aggregated prototype descriptor \(\mathbf{d}_{\mathrm{agg}}\) to refine the global descriptor through a gated residual injection. This is important because \(\mathbf{f}^{(g)}\) is produced by global aggregation and therefore primarily reflects spatially dominant evidence, whereas \(\mathbf{d}_{\mathrm{agg}}\) summarises complementary local structure preserved by SCRL. In forensic knife retrieval, such structure may include weak but class-separating cues distributed over limited spatial support and attenuated by global aggregation alone. Reciprocal injection allows prototype-informed local evidence to contribute to the final descriptor while preserving the robustness of the global representation.

\subsubsection{Global to Local Residual Projection}
Let \(\mathbf{f}^{(g)}\in\mathbb{R}^{K}\) be the global descriptor and let \(\mathbf{x}'_{h,w}\in\mathbb{R}^{K}\) denote the refined local feature at location \((h,w)\) in \(\mathbf{X}'\). CoRe-Net removes the component of each local vector aligned with \(\mathbf{f}^{(g)}\):
\begin{equation}
\mathbf{x}^{\mathrm{res}}_{h,w}
=
\mathbf{x}'_{h,w}
-
\frac{\langle \mathbf{f}^{(g)},\mathbf{x}'_{h,w}\rangle}{\|\mathbf{f}^{(g)}\|_2^2+\epsilon}\,\mathbf{f}^{(g)},
\label{eq:orth_local}
\end{equation}
where \(\epsilon>0\) is a small constant for numerical stability. Collecting \(\mathbf{x}^{\mathrm{res}}_{h,w}\) over all spatial locations yields the residual local feature map \(\mathbf{X}^{\mathrm{res}}\in\mathbb{R}^{K\times H\times W}\).

This operation enforces complementarity at the feature level by suppressing components of the refined local map that are already explained by the global descriptor, while preserving residual local structure not explicitly represented in \(\mathbf{f}^{(g)}\).

\subsubsection{Local to Global Reciprocal Injection}
The aggregated prototype descriptor \(\mathbf{d}_{\mathrm{agg}}\in\mathbb{R}^{K}\) is injected into the global descriptor through a learnable scalar gate:
\begin{equation}
\mathbf{f}^{(\mathrm{enh})}=\mathbf{f}^{(g)}+\eta\,\mathbf{d}_{\mathrm{agg}},
\label{eq:inj}
\end{equation}
where \(\eta>0\) is parameterised as \(\eta=\mathrm{softplus}(\eta_{\mathrm{raw}})\), with \(\eta_{\mathrm{raw}}\in\mathbb{R}\) being a learnable scalar parameter. We adopt an additive gated residual form for local-to-global injection because both \(\mathbf{f}^{(g)}\) and \(\mathbf{d}_{\mathrm{agg}}\) lie in the same feature space \(\mathbb{R}^{K}\). The role of \(\mathbf{d}_{\mathrm{agg}}\) is to provide a descriptor-level refinement of \(\mathbf{f}^{(g)}\). Residual addition preserves descriptor dimensionality, maintains the global branch as the anchor representation, and allows the contribution of the prototype-aggregated local summary to be modulated explicitly through a learnable gate. The softplus parameterisation ensures a strictly positive injection weight and stable optimisation.

\subsubsection{Fused Representation and Embedding}
We form a global feature map by replicating \(\mathbf{f}^{(\mathrm{enh})}\) across the spatial grid:
\begin{equation}
\mathbf{F}^{(\mathrm{enh})}_{:,h,w} = \mathbf{f}^{(\mathrm{enh})},
\quad \forall\, h\in\{1,\dots,H\},\, w\in\{1,\dots,W\},
\label{eq:broadcast}
\end{equation}
and concatenate it with the residual local map:
\begin{equation}
\mathbf{F} = \mathrm{Concat}\!\left(\mathbf{X}^{\mathrm{res}},\, \mathbf{F}^{(\mathrm{enh})}\right) \in \mathbb{R}^{2K\times H\times W}.
\label{eq:fused_tensor}
\end{equation}
The fused tensor \(\mathbf{F}\) is then globally pooled by adaptive average pooling and passed through a projection head comprising a linear layer, batch normalisation, and a PReLU activation to obtain the final embedding \(\mathbf{z}\in\mathbb{R}^{D}\), with \(D=512\) in our experiments. The resulting embedding is \(\ell_2\)-normalised at retrieval time and compared using cosine similarity.

\subsection{Training Objective}
CoRe-Net is trained end-to-end using a sub-center ArcFace classification objective and the prototype orthogonality regulariser:
\begin{equation}
\mathcal{L}=\mathcal{L}_{\mathrm{ArcFace}}+\lambda_{\mathrm{ortho}}\mathcal{L}_{\mathrm{ortho}},
\label{eq:loss}
\end{equation}
where \(\mathcal{L}_{\mathrm{ortho}}\) is computed as in \eqref{eq:ortho} and \(\lambda_{\mathrm{ortho}}\) controls the strength of prototype diversity enforcement. In this objective, \(\mathcal{L}_{\mathrm{ArcFace}}\) supervises the retrieval embedding, while \(\mathcal{L}_{\mathrm{ortho}}\) encourages diversity among prototypes and reduces collapse to redundant local representations.

\section{Results}
\label{sec:results}

We evaluate CoRe-Net on the KnifeHunter benchmark using single-descriptor cosine retrieval without query expansion, re-ranking, or geometric verification, following operational requirements for fast, deterministic, CPU-compatible deployment. We report backbone selection, comparison with state-of-the-art retrieval methods, ablations, parameter sensitivity, and qualitative analysis. All training and evaluation experiments are repeated five times with different random seeds, and mean performance is reported.

\subsection{Training and Evaluation Protocol}
\label{sec:training_protocol}

All images are resized to \(448 \times 448\) pixels. The default CoRe-Net configuration uses the eva02\_base\_patch14\_448 backbone (EVA02-Base), which produces patch tokens reshaped into a dense feature tensor \(\mathbf{X}\in\mathbb{R}^{768\times32\times32}\). CoRe-Net aggregates this tensor using the Weibull-based global branch, SCRL branch, and BDRF module, followed by global pooling and a projection head consisting of a linear layer, batch normalisation, and PReLU activation. The output embedding has dimension \(D=512\) and is used as the retrieval descriptor. During training, it is supervised with a sub-center ArcFace head using \(k=3\) sub-centers per class over 543 classes.

Training uses sub-center ArcFace with scale \(s=45\) and margin \(m=0.1\), together with the SCRL orthogonality regulariser weighted by \(\lambda_{\mathrm{ortho}}=1000\). All parameters, including the backbone, are optimised with Adam using an initial learning rate of \(1\times10^{-5}\). A step schedule with step size 4 and decay factor 0.8 is used. Models are trained for 30 epochs with batch size 64.

At test time, each query and database image is encoded once, \(\ell_2\)-normalised, and ranked by cosine similarity against precomputed database descriptors. We report mAP and mP@10 on the Medium and Hard protocols, with and without distractors, following Section~\ref{sec:dataset}. CoRe-Net contains 94M trainable parameters and requires 93.1 GMACs per forward pass, corresponding to approximately 186.2 GFLOPs under the \(2\times\)MAC convention. Descriptor extraction takes 20.8 ms per image on an NVIDIA RTX 3090Ti with batch size 1, excluding nearest-neighbour search.

\begin{table*}[t]
\centering
\caption{Backbone comparison using GeM pooling on KnifeHunter. We report mAP (\%) on Medium and Hard, with and without distractors (+desc). \emph{Backbone Params} denotes the number of parameters in the backbone network only (excluding GeM aggregation and the ArcFace head), rounded to the nearest million.}
\label{tab:baseline_backbones}
\setlength{\tabcolsep}{6pt}
\renewcommand{\arraystretch}{1.15}
\begin{tabular}{lccccc}
\toprule
& \multicolumn{2}{c}{\textbf{No distractors}} & \multicolumn{2}{c}{\textbf{+desc}} & \\
\cmidrule(lr){2-3}\cmidrule(lr){4-5}
\textbf{Backbone (GeM)}
& \textbf{Med. mAP} & \textbf{Hard mAP}
& \textbf{Med. mAP} & \textbf{Hard mAP}
& \textbf{Backbone Params} \\
\midrule
ConvNeXt-Base & 77.2 & 56.8 & 74.3 & 48.1 & 87M \\
SwinV2-Base   & 79.4 & 59.1 & 76.6 & 51.9 & 88M \\
EVA02-Base    & \textbf{81.1} & \textbf{61.1} & \textbf{78.3} & \textbf{54.2} & 88M \\
\bottomrule
\end{tabular}
\end{table*}

\begin{table*}[t]
\centering
\caption{Comparison of single-descriptor retrieval methods on KnifeHunter (EVA02-Base backbone). We report mAP and mP@10 (\%) on Medium and Hard protocols, with and without distractors (+desc).}
\label{tab:global_methods_map}
\setlength{\tabcolsep}{5pt}
\renewcommand{\arraystretch}{1.2}
\begin{tabular}{lccccccccc}
\toprule
& \multicolumn{4}{c}{\textbf{No distractors}} & \multicolumn{4}{c}{\textbf{+desc}} & \\
\cmidrule(lr){2-5}\cmidrule(lr){6-9}
\textbf{Method}
& \textbf{Med. mAP} & \textbf{Med. mP@10} & \textbf{Hard mAP} & \textbf{Hard mP@10}
& \textbf{Med. mAP} & \textbf{Med. mP@10} & \textbf{Hard mAP} & \textbf{Hard mP@10}
& \textbf{Params} \\
\midrule
GeM         & 81.1 & 79.8 & 61.1 & 58.6 & 78.3 & 77.2 & 54.2 & 50.9 & 92M \\
DOLG        & 83.4 & 82.2 & 63.4 & 61.1 & 80.6 & 79.5 & 56.4 & 53.6 & 117M \\
DELG        & 83.7 & 82.4 & 63.8 & 61.3 & 80.9 & 79.8 & 56.7 & 53.7 & 119M \\
TokenNet    & 82.8 & 81.5 & 62.8 & 60.6 & 80.1 & 78.9 & 55.9 & 52.6 & 120M \\
SuperGlobal & 84.5 & 83.4 & 64.5 & 62.1 & 81.7 & 80.6 & 57.6 & 54.5 & 92M \\
SENet       & 84.9 & 83.6 & 64.9 & 62.5 & 82.2 & 81.0 & 58.2 & 54.9 & 93M \\
\midrule
\textbf{CoRe-Net} & \textbf{88.0} & \textbf{86.7} & \textbf{70.2} & \textbf{67.9} & \textbf{85.1} & \textbf{83.8} & \textbf{64.9} & \textbf{61.5} & 94M \\
\bottomrule
\end{tabular}
\end{table*}

\subsection{Backbone Selection using GeM Pooling}
\label{sec:backbone_selection}

We quantify the effect of backbone architecture under a fixed and widely used aggregation baseline, GeM pooling~\cite{gem}. This controlled setting isolates backbone representation capacity from aggregation design and provides a principled basis for selecting a default backbone for subsequent experiments. All models are trained and tested on the KnifeHunter benchmark using the same dataset splits, input resolution, augmentation pipeline, optimiser, learning-rate schedule, batch size, and number of epochs.

Table~\ref{tab:baseline_backbones} shows that under identical GeM aggregation and optimisation settings, EVA02-Base yields a higher mAP than ConvNeXt-Base and SwinV2-Base across both difficulty tiers and under the +desc protocol. Based on this controlled comparison, we adopt EVA02-Base as the default backbone for all subsequent experiments unless stated otherwise.

\subsection{Comparison of Global Descriptor Methods}
\label{sec:global_methods}

We train and evaluate several state-of-the-art global and hybrid aggregation methods integrated with a common EVA02-Base backbone, including GeM, DOLG, DELG, TokenNet, SuperGlobal, SENet and the proposed CoRe-Net. To ensure a controlled comparison, all methods use the same KnifeHunter splits and training protocol described above while preserving task-specific loss design where applicable. Specifically, GeM, DOLG, TokenNet, SENet, SuperGlobal, and CoRe-Net are trained with a sub-centre ArcFace objective, while DELG is trained with its original multi-objective loss~\cite{delg}.

\paragraph{Quantitative results}
Table~\ref{tab:global_methods_map} reports mAP and mP@10. CoRe-Net achieves the best performance across all evaluation settings. On the Medium protocol, CoRe-Net improves mAP from 84.9\% to 88.0\% and mP@10 from 83.6\% to 86.7\% (best baseline, SENet). On the Hard protocol, CoRe-Net improves mAP from 64.9\% to 70.2\% and mP@10 from 62.5\% to 67.9\%. Under the +desc protocol, CoRe-Net improves Medium+desc mAP from 82.2\% to 85.1\% and mP@10 from 81.0\% to 83.8\%, and improves Hard+desc mAP from 58.2\% to 64.9\% and mP@10 from 54.9\% to 61.5\%. These results indicate improved discrimination under both difficulty tiers and increased robustness with distractor database.

CoRe-Net achieves these gains without a substantial increase in model size. CoRe-Net contains 94M parameters, comparable to GeM and SuperGlobal (92M) and SENet (93M), while outperforming larger architectures such as DOLG (117M), DELG (119M), and TokenNet (120M).

\paragraph{Analysis and discussion}
Local and global fusion approaches, such as DOLG and DELG, improve performance over GeM, indicating that incorporating local feature representations alongside global descriptors benefits fine-grained retrieval when global aggregation alone is insufficient. Advanced global aggregation strategies such as SuperGlobal and SENet provide further improvements by strengthening the single-descriptor representation with additional inductive bias. SuperGlobal enhances global pooling through region-aware context aggregation, while SENet injects structural self-similarity information that improves discrimination among visually similar instances. These effects are most apparent under the Hard and +desc protocols, where clutter, occlusion, and visually plausible distractors increase the likelihood of confusion under global pooling alone.

CoRe-Net improves over the baselines by combining robust global aggregation with structured local evidence modelling and reciprocal fusion. Weibull-based activation shaping reduces sensitivity to background-induced responses and high-magnitude outliers prior to pooling. SCRL organises high-variance local evidence into complementary prototype-based representations, while BDRF reduces redundancy between branches and injects prototype-derived local information into the global descriptor. This combination improves both overall ranking quality (mAP) and early precision (mP@10).

\begin{table*}[t]
\centering
\caption{Ablation study of CoRe-Net components. We report mAP (\%) on Medium and Hard protocols, with and without distractors.}
\label{tab:ablation_CoRe-Net}
\setlength{\tabcolsep}{6pt}
\renewcommand{\arraystretch}{1.15}
\begin{tabular}{lcc cc}
\toprule
& \multicolumn{2}{c}{\textbf{No distractors}} & \multicolumn{2}{c}{\textbf{+desc}} \\
\cmidrule(lr){2-3}\cmidrule(lr){4-5}
\textbf{Configuration} & \textbf{Med. mAP} & \textbf{Hard mAP} & \textbf{Med. mAP} & \textbf{Hard mAP} \\
\midrule
Global pooling: GeM (replace Weibull) & 86.1 & 67.8 & 83.2 & 61.3 \\
Local refinement: SGFR off (no variance mask) & 86.3 & 68.1 & 83.7 & 61.8 \\
Fusion: orthogonal only (no reciprocal injection), prototypes on & 85.3 & 66.1 & 82.5 & 60.1 \\
Local modeling: prototypes off (BDRF on, pooled-local injection) & 85.0 & 65.3 & 82.3 & 59.2 \\
\midrule
\textbf{CoRe-Net (all components)} & \textbf{88.0} & \textbf{70.2} & \textbf{85.1} & \textbf{64.9} \\
\bottomrule
\end{tabular}
\end{table*}

\subsection{Ablation Study of CoRe-Net Components}
\label{sec:ablation}

To quantify the contribution of each CoRe-Net component, we perform an ablation study with the EVA02-Base backbone. Starting from the full CoRe-Net configuration, we remove or replace one component at a time while keeping all other model components unchanged. We consider the following variants:
\begin{itemize}
    \item \emph{GeM instead of Weibull:} The Weibull-based global aggregation branch (Weibull shaping + global average pooling + power normalisation) is replaced by GeM pooling.
    \item \emph{No SGFR:} Saliency-Guided Feature Refinement is removed, and the local branch operates on the unmasked feature map.
    \item \emph{Orthogonal fusion instead of BDRF:} Fusion uses orthogonal projection of local features with respect to the global descriptor but disables the local-to-global injection.
    \item \emph{No prototypes (BDRF on, pooled-local injection):} Prototype-based local descriptor learning is removed. The BDRF pipeline is retained, but the injected vector is computed by global average pooling of the refined local map \(\mathbf{X}'\), i.e., \(\mathbf{d}_{\mathrm{pool}}=\mathrm{GAP}(\mathbf{X}')\in\mathbb{R}^{K}\) replaces \(\mathbf{d}_{\mathrm{agg}}\) in the local-to-global injection.
    \item \emph{Full CoRe-Net:} All components are enabled.
\end{itemize}

Replacing Weibull-based shaping with GeM reduces performance across all protocols. This indicates that activation shaping prior to pooling improves descriptor robustness under background clutter and high-magnitude activation outliers. Removing SGFR also degrades performance, with a 3.1 point reduction on Hard+desc, consistent with the role of the variance-based mask in suppressing regions with weak local variation and emphasising high-variance structure.

The fusion ablation shows that orthogonal fusion alone is insufficient; disabling the local-to-global injection reduces performance by 4.8 mAP points on Hard+desc. This indicates that, beyond redundancy suppression in the local branch, injecting structured local evidence into the global descriptor is beneficial when the global branch is affected by clutter, occlusion, or illumination artefacts.

Disabling prototypes yields the largest degradation, with a 5.7 point reduction on Hard+desc. This variant retains BDRF but replaces prototype aggregation with pooled-local injection, thereby isolating the contribution of prototype-based local modelling. Learnable prototypes provide a compact, structured summary of multiple local appearance patterns, whereas pooled local features provide a weaker and less discriminative injection signal. Overall, Table~\ref{tab:ablation_CoRe-Net} shows that each component contributes to retrieval accuracy and that the full combination is required to achieve the best performance.

\begin{table}[t]
\centering
\caption{Effect of the number of prototypes (\(N_p\)) in CoRe-Net.}
\label{tab:ablation_prototypes}
\setlength{\tabcolsep}{8pt}
\renewcommand{\arraystretch}{1.1}
\begin{tabular}{lcccc}
\toprule
\textbf{\(N_p\)} & \textbf{Medium} & \textbf{Hard} & \textbf{Medium+desc} & \textbf{Hard+desc} \\
\midrule
3  & 86.7 & 68.7 & 83.9 & 63.0 \\
4  & 87.9 & 69.5 & 85.0 & 64.6 \\
\textbf{5}  & \textbf{88.0} & \textbf{70.2} & \textbf{85.1} & \textbf{64.9} \\
6  & 86.2 & 68.1 & 83.2 & 61.7 \\
7  & 85.2 & 66.0 & 82.5 & 60.2 \\
\bottomrule
\end{tabular}
\end{table}

\subsection{Effect of the Number of Prototypes}
\label{sec:ablation_prototypes}

We evaluate the effect of the number of prototypes used in SCRL. The number of prototypes controls the granularity of the local summary produced by prototype aggregation.

Table~\ref{tab:ablation_prototypes} shows that the prototype count has a non-monotonic effect on retrieval performance. Increasing the number of prototypes from \(N_p=3\) to \(N_p=4\) and \(N_p=5\) improves accuracy, indicating that a small set of prototypes is beneficial for decomposing refined local structure into multiple complementary components. The strongest overall performance is obtained with \(N_p=5\), which achieves the best results on the Hard and Hard+desc protocols while remaining competitive on Medium and Medium+desc. In contrast, further increasing the number of prototypes to \(N_p=6\) and \(N_p=7\) degrades performance. This suggests that beyond a moderate number of components, prototype aggregation becomes overly fragmented, reducing the effectiveness of the resulting local summary for descriptor refinement and fusion.

\subsection{Effect of Orthogonality Regularisation Weight}
\label{sec:ablation_lambda}

We analyse the sensitivity of CoRe-Net to the orthogonality regularisation weight \(\lambda_{\text{ortho}}\) used to decorrelate prototypes in SCRL. All other model components and training settings are fixed, and only \(\lambda_{\text{ortho}}\) is varied. Setting \(\lambda_{\text{ortho}}=0\) reduces performance across all protocols (Table~\ref{tab:ablation_lambda}), indicating that unconstrained prototypes become correlated and provide a less informative local summary. Increasing \(\lambda_{\text{ortho}}\) improves performance, with the best results at \(\lambda_{\text{ortho}}=1000\). Further increasing to 1500 degrades performance, indicating that overly strong decorrelation restricts the prototype space and reduces adaptability to fine-grained variations.

\begin{table}[t]
\centering
\caption{Effect of the orthogonality regularisation weight \(\lambda_{\mathrm{ortho}}\).}
\label{tab:ablation_lambda}
\setlength{\tabcolsep}{8pt}
\renewcommand{\arraystretch}{1.1}
\begin{tabular}{lcccc}
\toprule
\textbf{\(\lambda_{\text{ortho}}\)} & \textbf{Medium} & \textbf{Hard} & \textbf{Medium+desc} & \textbf{Hard+desc} \\
\midrule
0     & 85.1 & 66.1 & 82.3 & 60.1 \\
500   & 86.6 & 68.3 & 83.8 & 63.0 \\
\textbf{1000}  & \textbf{88.0} & \textbf{70.2} & \textbf{85.1} & \textbf{64.9} \\
1500  & 87.9 & 69.8 & 85.0 & 64.3 \\
\bottomrule
\end{tabular}
\end{table}

\begin{figure*}[!t]
    \centering
    \includegraphics[width=\textwidth]{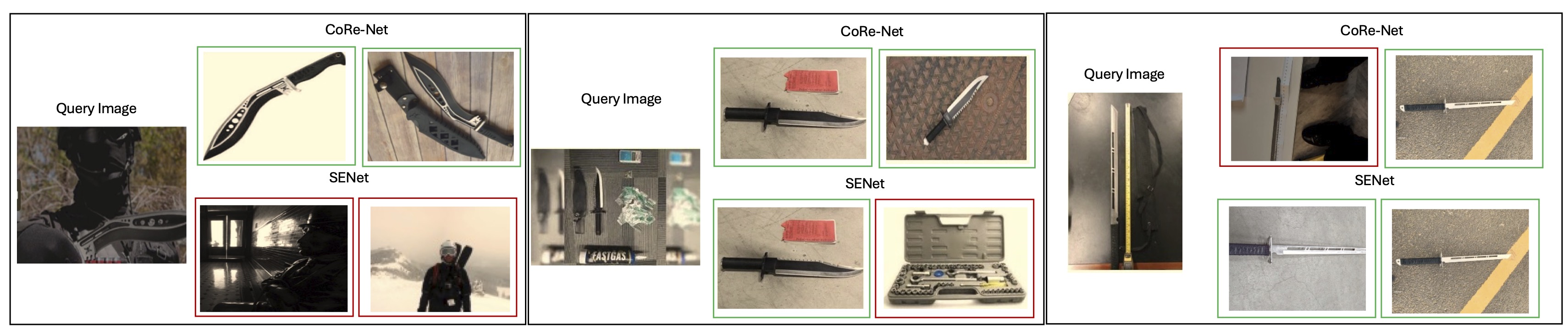}
    \caption{Qualitative comparison of retrieval results under the distractor protocol. Each row shows a query from the Hard test set. The retrieval database contains the training set augmented with distractors. For each query, we show the top-2 retrieved results produced by CoRe-Net and SENet with an EVA02-Base backbone. Green borders denote correct matches and red borders denote incorrect matches.}
    \label{fig:retrieval_vis}
\end{figure*}

\subsection{Qualitative Retrieval Analysis}
\label{sec:qualitative}

Figure~\ref{fig:retrieval_vis} shows qualitative retrieval results under the distractor protocol for Hard queries. Both methods use the EVA02-Base backbone and differ only in the aggregation module. CoRe-Net retrieves the correct knife instance more consistently than SENet in these examples, particularly under background clutter, evidence markers, challenging illumination, or limited visible knife detail. This visual comparison is consistent with the quantitative results in Table~\ref{tab:global_methods_map}, where CoRe-Net improves over SENet across the Hard and +desc protocols. The final query illustrates a remaining failure mode: when discriminative knife evidence is severely limited and strong co-occurring linear structures are present, visually plausible incorrect matches may still be ranked highly. Detailed retrieval analysis and failure cases are provided in Supplementary Sections B and C.

\subsection{Deployment, External Validation, and Policy Impact}
\label{sec:deployment}

KnifeHunter was deployed on Amazon Web Services cloud infrastructure to provide secure access for UK police forces. Officers upload operational knife images through a web portal or specialised mobile application; the back-end performs descriptor extraction and similarity retrieval, then automatically generates a structured intelligence report containing the matched weapon characteristics. The secure web interface also supports database management and analytical dashboards for monitoring weapon submissions, classifications, and geographic patterns, as illustrated in Supplementary Fig.~4.

KnifeHunter was first validated during an operational trial as part of \textit{Operation Sceptre} in May 2023 and was subsequently evaluated during further \textit{Operation Sceptre} deployments in 2024 and 2025. Across field queries collected from these operational deployments, the system achieved 99.2\% mP@1, demonstrating reliable top-ranked retrieval in frontline policing workflows. It remains in active use, with police forces from 22 UK regions integrating it into operational practice. The reference database is continually enriched through annotation by industry and forensic experts, improving the accuracy and evidentiary value of system outputs.

Beyond instance retrieval, KnifeHunter supports structured intelligence gathering on weapon typology, retail availability, and online sales patterns. Data collected by the system contributed to the independent end-to-end review of knife sales in the UK~\cite{clayman2025knife}, supporting recommendations on more consistent offence recording, potential retailer registration, evidence for action against non-compliant overseas sellers, and intelligence-led enforcement by police forces, Trading Standards, and Ofcom. By automating visual matching and cataloguing, KnifeHunter reduces manual administrative burden while enabling scalable cross-case analysis and policy-relevant evidence generation. More detail on UK Gov policy is presented in Supplementary section E.

\section{Conclusion}
\label{sec:conclusion}

This paper introduced KnifeHunter, an operationally deployed system for fine-grained knife image retrieval in law enforcement. We presented the KnifeHunter dataset, comprising 25{,}843 images across 543 knife classes from police evidence repositories, retail catalogues, and border-force seizures, with Medium/Hard protocols and large-scale distractor evaluation. We also proposed CoRe-Net, a compact single-descriptor retrieval architecture that combines SCRL for prototype-based modelling of local discriminative evidence with BDRF for complementary global-local descriptor formation.

Experiments show that CoRe-Net consistently outperforms strong global and hybrid baselines across both difficulty tiers. Under the distractor protocol, CoRe-Net achieves 85.1\% mAP and 83.8\% mP@10 on Medium, and 64.9\% mAP and 61.5\% mP@10 on Hard, demonstrating improved robustness in large-scale retrieval. KnifeHunter has also been deployed within UK policing infrastructure, supporting operational image search, structured reporting, intelligence analysis, and policy activities related to knife sales.

Future work will focus on improving robustness under severe occlusion and specular saturation, incorporating metadata and multimodal cues for constrained search, and extending evaluation to open-set conditions involving unseen or modified knife classes.

\section*{Acknowledgments}
We acknowledge the National Police Chiefs' Council Knife Crime Working Group for supporting the deployment of KnifeHunter across UK police forces.

\section*{Funding}
This work was supported by the Police STAR Fund project.

\section*{Data Availability}
The dataset generated and analysed during this study can be obtained from  
\url{https://doi.org/10.5281/zenodo.19210078}

\vfill

\end{document}